\documentclass[letterpaper, 10 pt, conference]{ieeeconf}  
\usepackage{subcaption}
\usepackage{graphicx}
\usepackage{float}
\usepackage{amsmath}
\usepackage{amssymb}
\usepackage{xcolor}
\usepackage{hyperref}
\usepackage{epsfig,subcaption}
\usepackage[linesnumbered,ruled,vlined]{algorithm2e}
\usepackage{footmisc}
\usepackage[utf8]{inputenc}
\usepackage{algorithmic}
\usepackage{tabularray}
\usepackage{setspace}
\usepackage[noadjust]{cite}
\usepackage{url}
\usepackage{multirow}
\newcommand{\any}{AnyTraverse\ }

\IEEEoverridecommandlockouts                    

\title{\LARGE \bf
AnyTraverse: An off-road traversability framework with VLM and human operator in the loop
}

\author{Sattwik Sahu$^{1}$, Agamdeep Singh$^{1}$, Karthik Nambiar$^{1}$, Srikanth Saripalli$^{2}$, and P.B. Sujit$^{1}$
\thanks{$^{1}$Sattwik Sahu, Agamdeep Singh, Karthik Nambiar and P.B. Sujit is with Department of Electrical Engineering and Computer Science,
        Indian Institute of Science Education and Research Bhopal, India
        {\tt\small \{sattwik21, agamdeep20, karthik23, sujit\}@iiserb.ac.in}}%
\thanks{$^{2}$Srikanth Saripalli is with the Department of Mechanical Engineering, Texas A\&M University, USA
        {\tt\small ssaripalli@tamu.edu}}%
}

\begin{document}

\maketitle
\thispagestyle{empty}
\pagestyle{empty}

\begin{abstract}
 Off-road traversability segmentation enables autonomous navigation with applications in search-and-rescue, military operations, wildlife exploration, and agriculture. Current frameworks struggle due to significant variations in unstructured environments and uncertain scene changes, and are not adaptive to be used for different robot types. We present AnyTraverse, a framework combining natural language-based prompts with human-operator assistance to determine navigable regions for diverse robotic vehicles. The system segments scenes for a given set of prompts and calls the operator only when encountering previously unexplored scenery or unknown class not part of the prompt in its region-of-interest, thus reducing active supervision load while adapting to varying outdoor scenes. Our zero-shot learning approach eliminates the need for extensive data collection or retraining. Our experimental validation includes testing on RELLIS-3D, Freiburg Forest, and RUGD datasets and demonstrate real-world deployment on multiple robot platforms. The results show that AnyTraverse performs better than GA-NAV and Off-seg while offering a vehicle-agnostic approach to off-road traversability that balances automation with targeted human supervision.
 \end{abstract}

\section{Introduction}
\begin{figure}[h]
    \centering
    \begin{subfigure}[b]{0.3\columnwidth}
        \centering
        \includegraphics[width=\textwidth, height=1in]{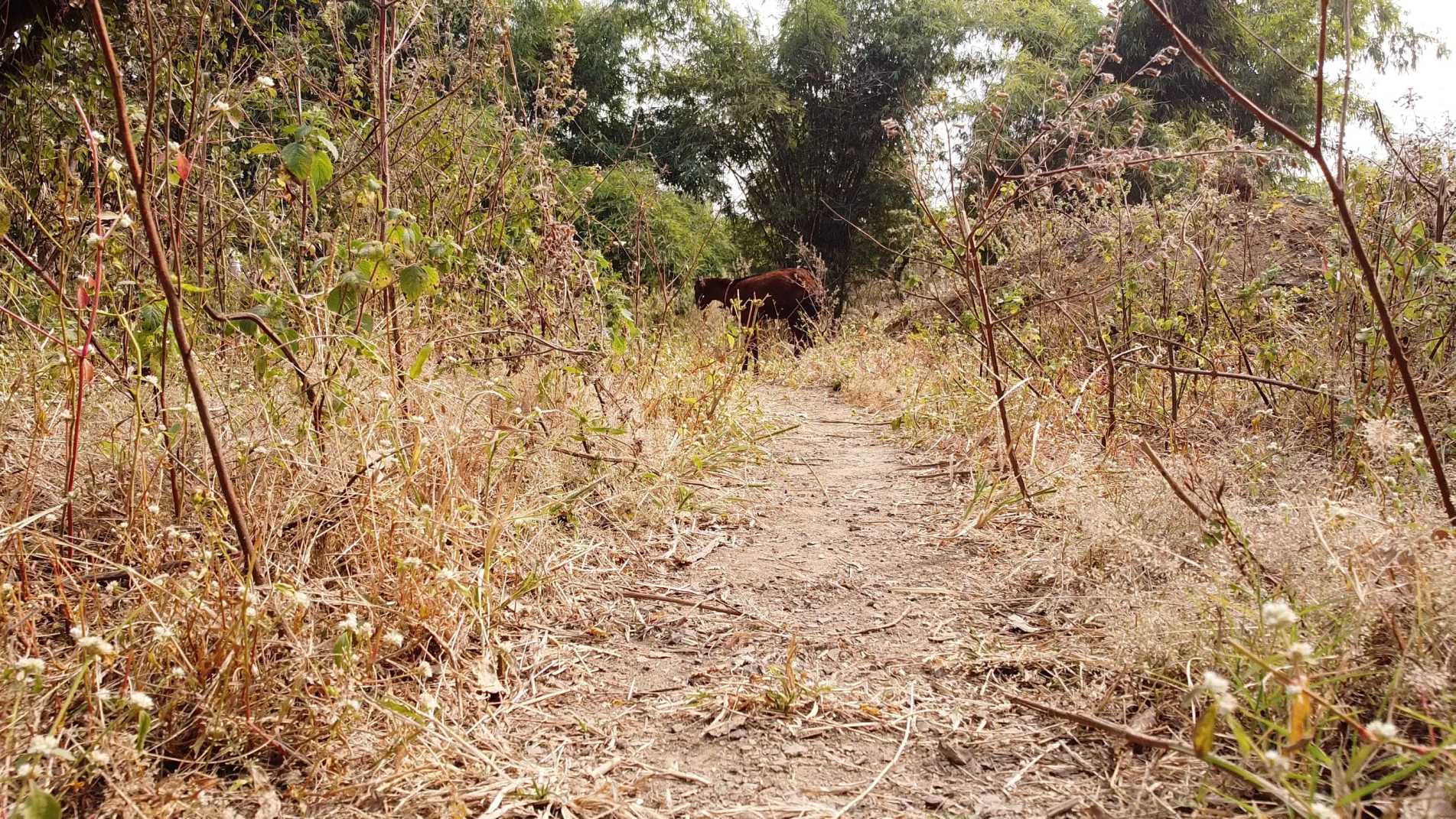}
        \caption{}
    \end{subfigure}
    \begin{subfigure}[b]{0.3\columnwidth}
        \centering
        \includegraphics[width=\textwidth, height=1in]{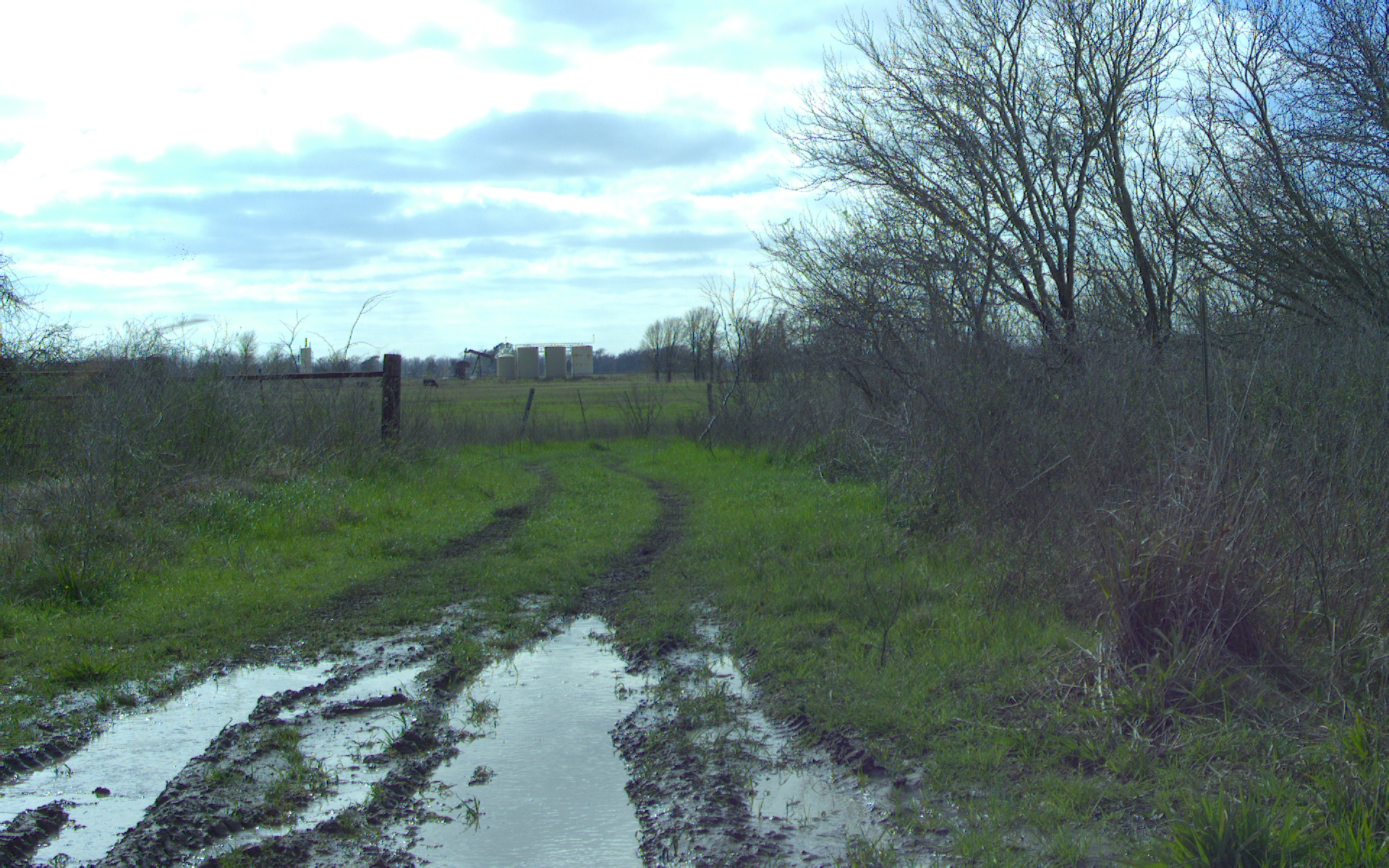}
        \caption{}
    \end{subfigure}
    \begin{subfigure}[b]{0.3\columnwidth}
        \centering
        \includegraphics[width=\textwidth, height=1in]{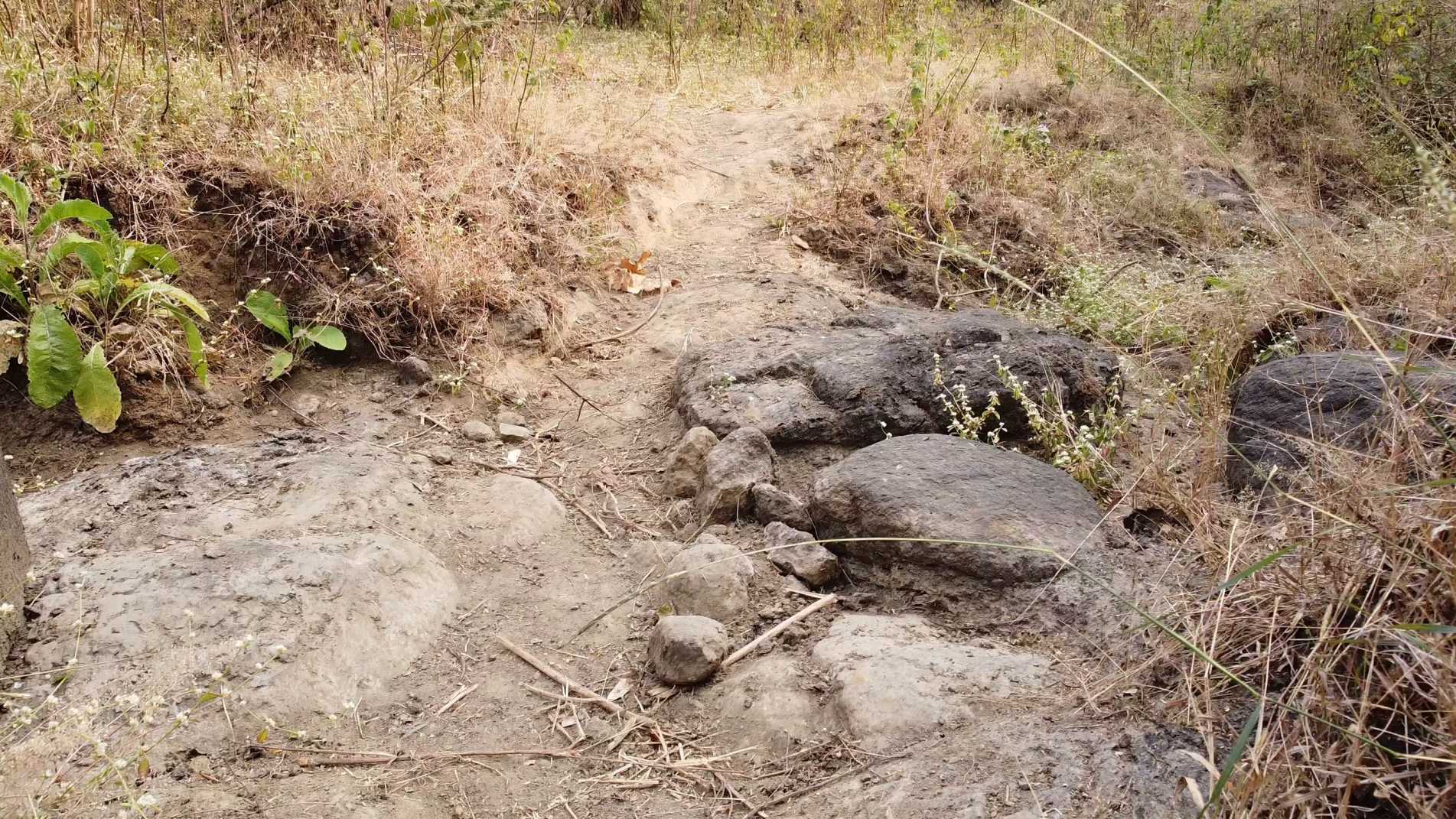}
        \caption{}
    \end{subfigure}
    \caption{Diverse off-road navigation scenarios 
    (a) Dense Vegetation and narrow trail (b) Puddles and dirt (c) Rocky terrain.}
    \label{fig:scenes}
\end{figure}

Navigation in off-road environments requires accurate classification of traversable regions from camera images. Unlike structured road environments, off-road scenes exhibit significant variations (\autoref{fig:scenes}), making it difficult to develop comprehensive models due to limited datasets capturing diverse terrain conditions. Furthermore, traversability definitions vary by vehicle type—regions navigable by a quadruped may be impassable for a wheeled rover. Traditional learning-based approaches cannot generalize effectively to novel environments and are typically optimized for specific robots. AnyTraverse addresses three key challenges: identifying traversable regions in highly variable unstructured environments, adapting to different vehicle types with varying traversability definitions, and efficiently leveraging human operator expertise while minimizing supervision load.

Prior work in off-road semantic segmentation includes deep learning approaches such as OffSeg~\cite{offseg}, which introduced a framework using BiSeNet \cite{bisenet} and HRNet \cite{hrnet} to classify terrains, and GA-Nav~\cite{ga-nav}, which employed a group-wise attention mechanism to detect navigable areas. These methods rely on fixed navigation systems with pre-defined classes and struggle with environmental changes or novel obstacles. Vision-Language Models (VLMs) address these limitations through few-shot or zero-shot capabilities \cite{behav, tgs}. CLIPSeg~\cite{clipseg} generates binary segmentation maps using text prompts, while LSeg~\cite{lseg} and DenseClip~\cite{dense-clip} adapt CLIP \cite{clip} for pixel-level segmentation. For navigation specifically, CoNVOI~\cite{convoi} proposed a VLM-based system for context-aware autonomous navigation, while TGS~\cite{tgs} leveraged VLMs for trajectory generation in outdoor environments. Multiple datasets support off-road segmentation research, including DeepScene~\cite{deepscene}, RUGD~\cite{rugd}, RELLIS-3D~\cite{rellis3d}.

Our contributions include a novel off-road traversability framework using CLIPSeg as a backbone VLM with selective human operator intervention, a region-of-interest and scene similarity mechanism for determining when human input is required, comprehensive evaluation across multiple datasets and real-world robot platforms, and detailed performance analysis with operator workload quantification.
\section{Technical Approach}

\subsection{Attention Maps for Prompts} {
    \any operates as an operator-assisted vision-language model framework for determining traversable regions in off-road environments (Fig.~\ref{fig:pipeline}). The system comprises a VLM-based segmentation pipeline that generates binary traversability masks using natural language prompts and a traversability evaluator that detects significant scene changes and unknown objects in the robot's path.
    
    Given an input RGB image at time $t$, $I_t \in \mathbb{R}^{H\times W\times C}$ and a list $\tau$ of prompt-weight pairs provided by the human operator: $\tau = \{(\pi_1,w_1),...,(\pi_k,w_k)\}$, where $\pi_n$ are natural language prompts describing terrain elements (e.g., ``grass,'' ``rocks,'' ``water'') and $w_n \in [-1,1]$ are their corresponding traversability weights ($1$ for fully traversable, $-1$ for non-traversable), the VLM model (\textit{e.g.} CLIPSeg~\cite{clipseg}), $\mathcal{M}_\text{map}$ generates attention maps for each prompt: $m_n = \mathcal{M}_\text{map}(I_t, \pi_n)$, where $m_n \in [0,1]^{H\times W}$ is the attention map for prompt $\pi_n$.
    
    \begin{figure}[t]
        \centering
        \includegraphics[width=\linewidth]{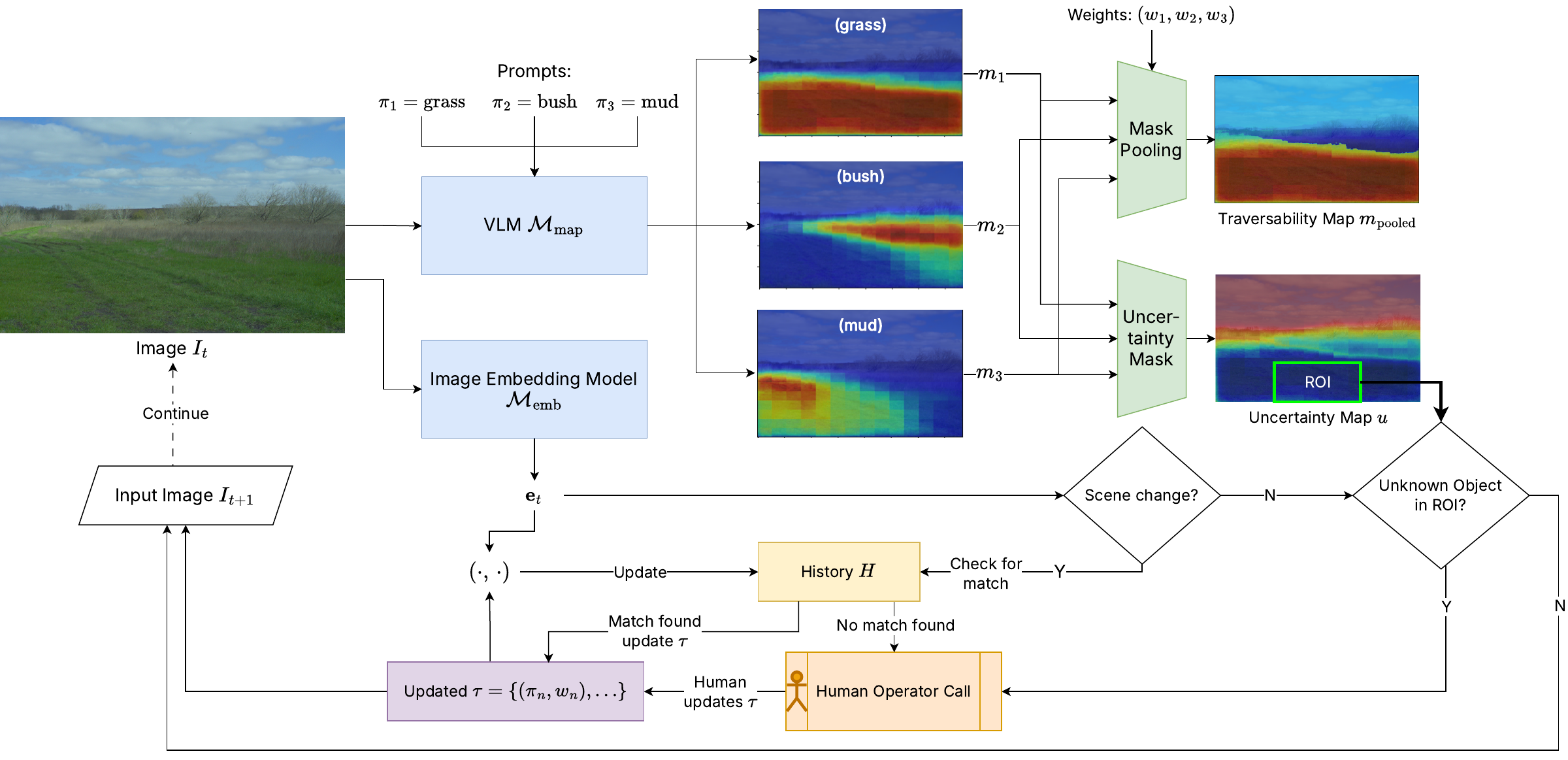}
        \caption{\any starts with an image $I_t$, from which masks $m_n$ are created given prompts $\pi_n$ from user. These masks are pooled using mask pooling to create $m_\text{pooled}$. The human operator is called when either a scene change is detected (using the embedding $\mathbf{e}_t$ of $I_t$ from image embedding model $\mathcal{M}_\text{emb}$) or the uncertainty in the ROI is higher than threshold, \textit{i.e.}, an unknown object is detected in the ROI of the vehicle. After a human operator call modifies the traversability preferences $\tau$, the image embedding $\mathbf{e}_t$ and the updated $\tau$ are stored in the history $H$. In case of a scene change, the pipeline tries to find a scene similar to the current one in the history $H$ and updates the $\tau$ using the traversability preferences for the matching scene. If no match is found, it proceeds to call the human operator.}
        \label{fig:pipeline}
    \end{figure}
}

\subsection{Weighted Pooling of Masks for Traversability Segmentation} {
    We introduce a novel weighted mask pooling algorithm that combines multiple attention maps, $m_i$ into a single traversability map, $m_\text{pooled} \in [0,\, 1]^{H \times W}$, described in Algorithm~\ref{alg:weighted_max_pooling}. The pooled mask, $m_{\text{pooled}}$, is first set to an empty $H \times W$ array, on line~\ref{alg:wmp_init}. The weighted masks, $\mu_n$, are calculated on line~\ref{alg:wmp_weight}, by taking the product of each mask, $m_n$, and its corresponding weight, $w_n$. For each element position, the index $\eta$ where the absolute value of the weighted mask, $|\mu_\eta|$, is maximum is identified on~line~\ref{alg:wmp_max-abs}, and $\mu_\eta$ is then assigned to $m_{\text{pooled}}$, at that position, on line~\ref{alg:wmp_assign}.
    
    \begin{algorithm}[t]
    \caption{Weighted Max Pooling Algorithm}
    \label{alg:weighted_max_pooling}
    
    \SetAlgoLined
        \KwIn {
            $\{(m_1, w_1), \dots, (m_k, w_k)\}$, mask and weight pairs
        }
        \KwOut {
            Pooled mask $m_{\text{pooled}} \in [0, 1]^{H \times W}$
        }
        \vspace{0.1cm}
        Initialize $m_{\text{pooled}} \gets \mathbf{0}^{H \times W}$\; \label{alg:wmp_init}
        Calculate $\mu_n = w_n m_n\ \forall\ n \in \{1, \dots, k\}$\; \label{alg:wmp_weight}
        \For{$(i, j) \in \{1, \dots, H\} \times \{1, \dots, W\}$} {
            $\eta \gets \underset{n \in {\{1, \dots, k\}}}{\arg\max}{|\mu_n[i, j]|}$\; \label{alg:wmp_max-abs}
            $m_{\text{pooled}}[i, j] \gets \mu_\eta$\; \label{alg:wmp_assign}
        }
        \KwRet $m_{\text{pooled}}$
    \end{algorithm}
}

\subsection{Region of Interest (ROI)} {
    \label{sec:method_roi}
    The region of interest (ROI) is the area in the immediate surroundings of the vehicle where it is most likely to traverse in the near future. Different vehicles vary in size and characteristic speeds at which they travel, which affects the size and shape of the ROI. \autoref{fig:vehicle-roi} shows how different vehicles would require different ROIs in the image $I_t$. This ROI is proposed by the human operator and is set before starting the \any pipeline. \any is designed to call the human operator whenever an unknown object enters the ROI, as it may be non-traversable and may risk failing the mission or damaging the vehicle. \autoref{sec:unk-obj-det} explains how the unknown object is detected. It must be optimized per vehicle to detect unknown objects while avoiding overloading the human operator with frequent calls.

    \begin{figure}[t]
        \centering
        \includegraphics[width=0.5\linewidth]{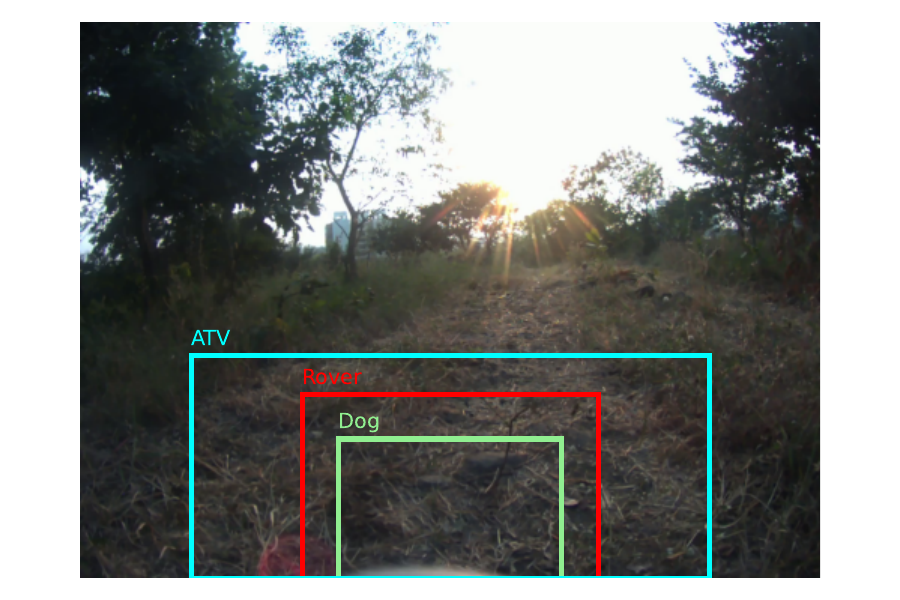}
        \caption{ROIs for different vehicles such as a quadruped (green), a rover (red) and an all-terrain vehicle \textit{aka} ATV (blue)}
        \label{fig:vehicle-roi}
    \end{figure}
}

\subsection{Uncertainty Maps to Detect Unknown Objects} {
    \label{sec:unk-obj-det}
    \any only has information on parts of the image which match any of the prompts $\pi_n$ in the image $I_t$. To detect unknown objects, we use \autoref{eq:invmax-proba-unc} to obtain an ``uncertainty'' metric for each pixel in the image, producing an uncertainty map $m_\text{unc}$. By performing an average over all the pixels in the ROI, we obtain an uncertainty score on the ROI, $u_\text{ROI} = \mathbb{E}_\text{ROI}[m_\text{unc}]$. An unknown object is said to be detected in the ROI when $u_\text{ROI} > \theta_\text{ROI}$, where $\theta_\text{ROI}$ is the threshold of ROI uncertainty score chosen by the human operator.
    \begin{equation}
        \label{eq:invmax-proba-unc}
        m_\text{unc}[i, j] = 1 - \left[\underset{n \in \{1, 2, \dots, k\}}{\max}{m_n[i, j]}\right]
    \end{equation}
}

\subsection{History} {
    \any maintains a history of human operator calls, containing the corresponding image embeddings (\textit{e.g.} using CLIP~\cite{clip}) $\mathbf{e}_c = \mathcal{M}_\text{emb}(I_c)$ and the updated traversability preferences $\tau_c$ provided by the human operator for that frame, where $c$ is a time at which the human operator was called. We denote the history by $H = \{(\mathbf{e}_{c_1}, \tau_{c_1}), (\mathbf{e}_{c_2}, \tau_{c_2}), \dots\}$, where $c_1, c_2, \dots$ are the times at which the human operator was called.
}

\subsection{Human Operator Calls (HOC)} {
    The scene is said to have changed when the current scene frame embedding $\mathbf{e}_t~=~\mathcal{M}_\text{emb}(I_t)$ is sufficiently different from a reference scene embedding, $\mathbf{e}^*$. At every time $t$, we check the similarity between $I_t$ and $I^*$ by $s_t = \sigma(\mathbf{e}_t, \mathbf{e}^*)$ where $\sigma(\cdot, \cdot)$ is the cosine similarity function. If $s_t < \theta_\text{scene}$, then we proceed to check the history $H$ for a similar frame encountered before, and update the traversability preferences accordingly. Here, $\theta_\text{scene}$ is the scene similarity threshold. A match in the history is found using \eqref{eq:match-mem}.
    \begin{equation}
        \label{eq:match-mem}
        (\mathbf{e}_{\text{match}}, \tau_{\text{match}}) = \underset{(\mathbf{e}', \tau') \in H}{\arg\min}\ {\sigma(\mathbf{e}_t, \mathbf{e}')} \\
    \end{equation}
    
    If $\sigma(\mathbf{e}_t, \mathbf{e}_\text{match}) \ge \theta_\text{scene}$, then it is a valid match and we proceed to update the traversability preferences by  If not, then this it a scene never encountered before and we proceed to call the human operator. The human operator provides a partial update for the traversability preferences, $\tau_\text{human}$, and the $\tau$ is then updated as $\tau \gets \nu(\tau, \tau_\text{human})$ (See \eqref{eq:tau-update} and \eqref{eq:merge-trav}). This history checking mechanism reduces load on the human operator by helping the robot adapt to a change in scene by using previous experience (within an episode).
    \begin{equation}
        \label{eq:tau-update}
        \tau \gets \begin{cases}
            \nu(\tau, \tau_\text{match}) \quad \text{if}\ \sigma(\mathbf{e}_t, \mathbf{e}_\text{match}) \ge \theta_\text{scene} \\
            \nu(\tau, \tau_\text{human}) \quad \text{otherwise} \\
        \end{cases}
    \end{equation}
    \begin{equation}
        \label{eq:merge-trav}
        \nu(\tau_1, \tau_2) = \tau_2 \cup \{(\pi', w') \in \tau_1 \mid (\pi', \hat{w}) \notin \tau_2 \quad \forall\ \hat{w} \in [-1, 1] \}
    \end{equation}

    Similarly, when an unknown object s detected in the ROI, the human operator is directly called and the traversability preferences $\tau$ are updated to $\nu(\tau, \tau_\text{human})$ where $\tau_\text{human}$ is the partial update given by the human like before.

    For every HOC, after updating $\tau$, we also update the reference scene embedding $\mathbf{e}^* \gets \mathbf{e}_t$ and the history $H \gets H \cup \{(\mathbf{e}_t, \tau)\}$ 
}

\section{Experiments Completed}

\subsection{Segmentation} 
    We evaluated AnyTraverse on multiple standard off-road datasets: RELLIS-3D (challenging off-road environments), RUGD (featuring diverse terrain types), and DeepScene (forested environments with natural obstacles). For each dataset, we created robot-specific prompt sets with appropriate traversability weights as shown in \autoref{tab:comparative}. To understand how AnyTraverse compares with existing methods, we benchmark against state-of-the-art off-road semantic segmentation techniques. \autoref{tab:comparative} presents a comparative analysis across the RELLIS-3D and RUGD datasets. While AnyTraverse is not the top performer in every category, it demonstrates competitive performance while offering the added benefits of adaptability and selective human intervention.

\subsection{Human Operator Call Analysis} {
    We conducted experiments to quantify operator intervention frequency across different datasets, tracking calls due to low ROI traversability confidence, calls due to unseen environment detection, and total intervention requests. \autoref{fig:hoc-plots} shows how operator calls accumulate over time with different similarity thresholds, $\theta_{\text{sim}}$. The results demonstrate the system's ability to learn from experience, requiring fewer interventions as it builds environmental familiarity.

\begin{table*}
    \centering
    \small
    \resizebox{\textwidth}{!}{%
    \begin{tabular}{|l|c|c|c|p{4.8cm}|}
        \hline
        \textbf{Dataset} & \textbf{OffSeg (Finetuned)} & \textbf{GA-Nav} & \textbf{AnyTraverse} & \textbf{Primary Prompts (Weight)} \\
        \hline
        RELLIS-3D\cite{rellis3d} & \textbf{0.866} & 0.744 & \underline{0.815} & grass (+1), bush (-1), dirt (+1) \\
        \hline
        RUGD\cite{rugd} & 0.845 & \textbf{0.891} & 0.834 & grass (+1), bush (-1), gravel (+1), water (-1) \\
        \hline
        DeepScene\cite{deepscene} & -- & -- & \textbf{0.852} & grass (+1), bush (-1), path (+1), water (-1) \\
        \hline
    \end{tabular}%
    }
    \caption{Comparative MIoU performance across datasets with key prompt configurations. Best scores are in bold, second-best are underlined.}
    \label{tab:comparative}
\end{table*}

    \begin{figure}
        \centering
        \begin{subfigure}[b]{0.24\columnwidth}
            \centering
            \includegraphics[width=\textwidth]{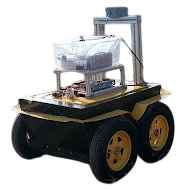}
            \caption{}
            \label{fig:robot-hound}
        \end{subfigure}
        \begin{subfigure}[b]{0.24\columnwidth}
            \centering
            \includegraphics[width=\textwidth]{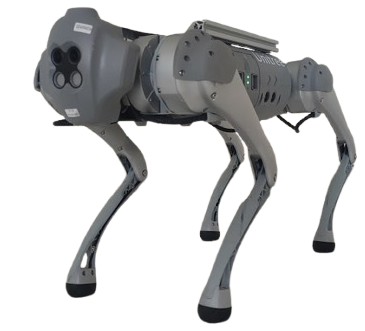}
            \caption{}
            \label{fig:robot-kombai}
        \end{subfigure}
        \begin{subfigure}[b]{0.24\columnwidth}
            \centering
            \includegraphics[width=\textwidth]{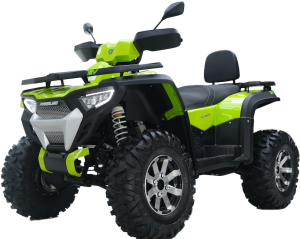}
            \caption{}
            \label{fig:robot-atv}
        \end{subfigure}
        \begin{subfigure}[b]{0.24\columnwidth}
            \centering
            \includegraphics[width=\textwidth]{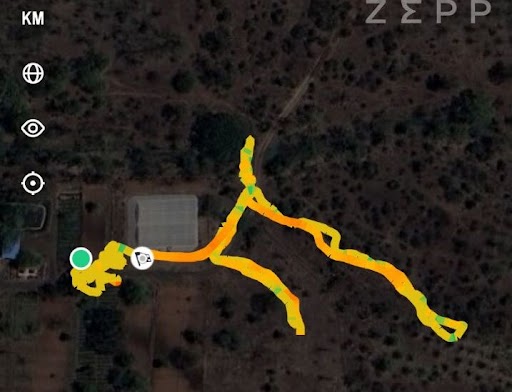}
            \caption{}
            \label{fig:robot-path}
        \end{subfigure}
        \caption{(a) 4 wheeled ground vehicle. (b) Unitree-GO1 quadrapedal robot. (c) All-Terrain Vehicle (ATV) (d) Path traversed for experiments in Forest Nursery(Bhopal, India).}
        \label{fig:robots-and-path}
    \end{figure}
    
    We deployed AnyTraverse on two robot platforms: a quadruped robot $\mathcal{R}_1$ (Unitree Go1) (\autoref{fig:robot-kombai}) and a custom-designed wheeled rover $\mathcal{R}_2$ (\autoref{fig:robot-hound}). For each platform, we developed specific prompt sets to account for their different locomotion capabilities. $\mathcal{R}_1$ allowed for traversal of rougher terrain like small rocks, while $\mathcal{R}_2$ was more conservative for these obstacles. Both platforms were tested in various environments, including forest trails, open fields, and hillside terrain. The forest trail was in a Forest Nursery in Sehore, Bhopal, which had various offroad terrain features such as gravel, streams, grass, dense bushes and animals. The evaluation in this environment is shown in \autoref{fig:hoc-ours}.
    \begin{figure}[ht]
        \centering
        \begin{subfigure}[b]{0.49\columnwidth}
            \centering
            \includegraphics[width=\textwidth]{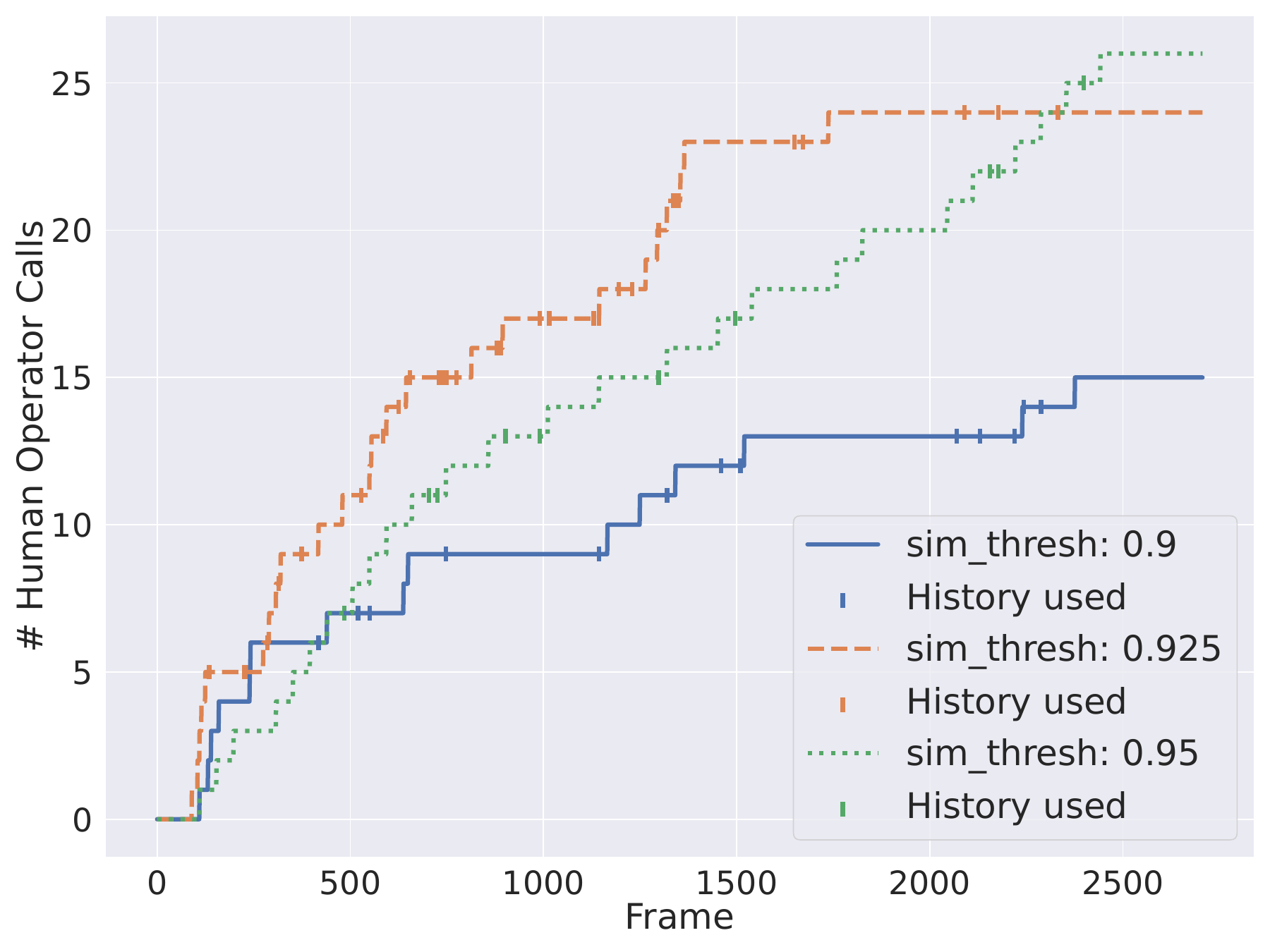}
            \caption{RELLIS-3D Episode 3}
            \label{fig:hoc-rellis}
        \end{subfigure}
        \begin{subfigure}[b]{0.49\columnwidth}
            \centering
            \includegraphics[width=\textwidth]{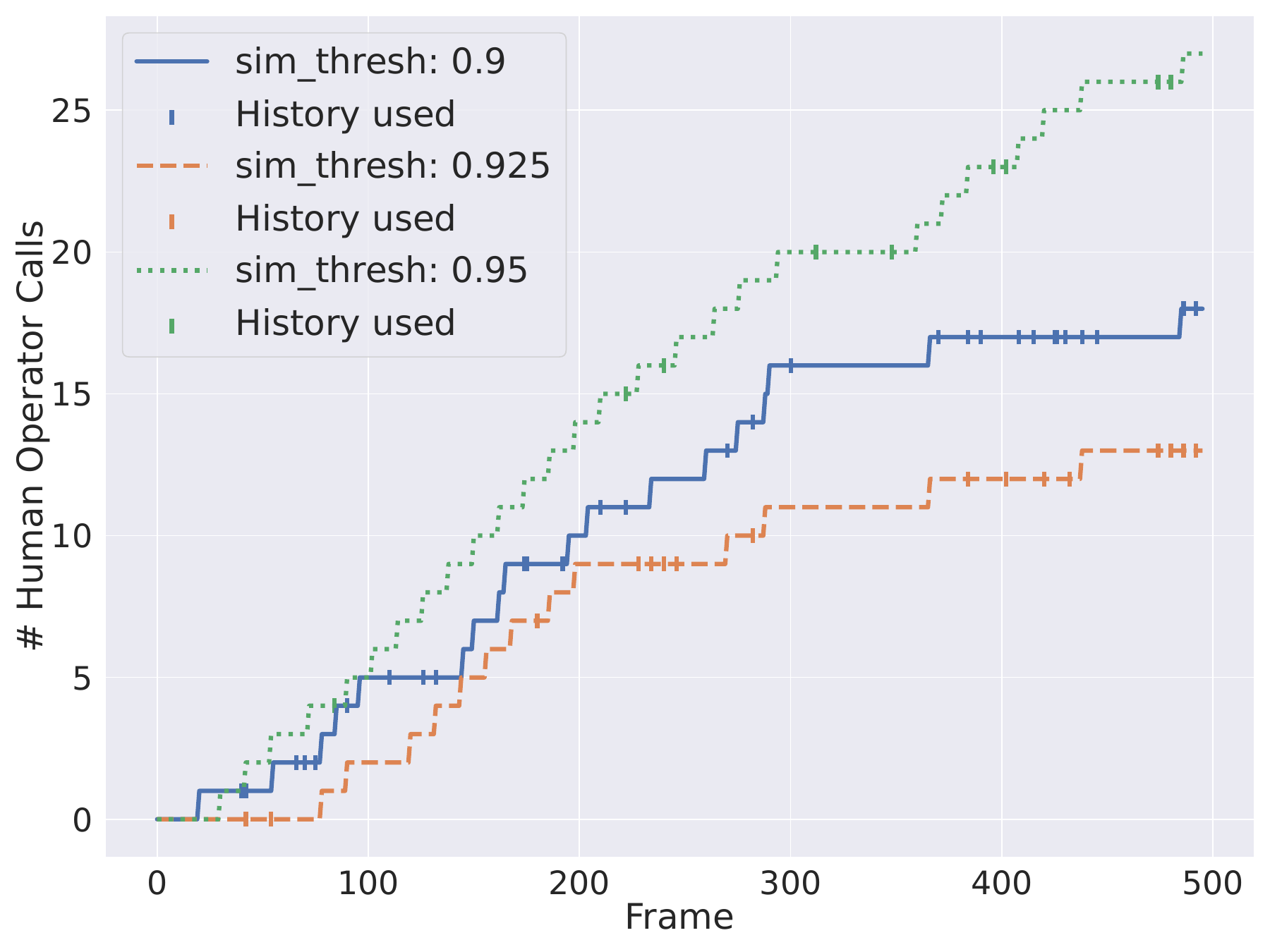}
            \caption{RUGD dataset}
            \label{fig:hoc-rugd}
        \end{subfigure}
        \begin{subfigure}[b]{0.49\columnwidth}
            \centering
            \includegraphics[width=\textwidth]{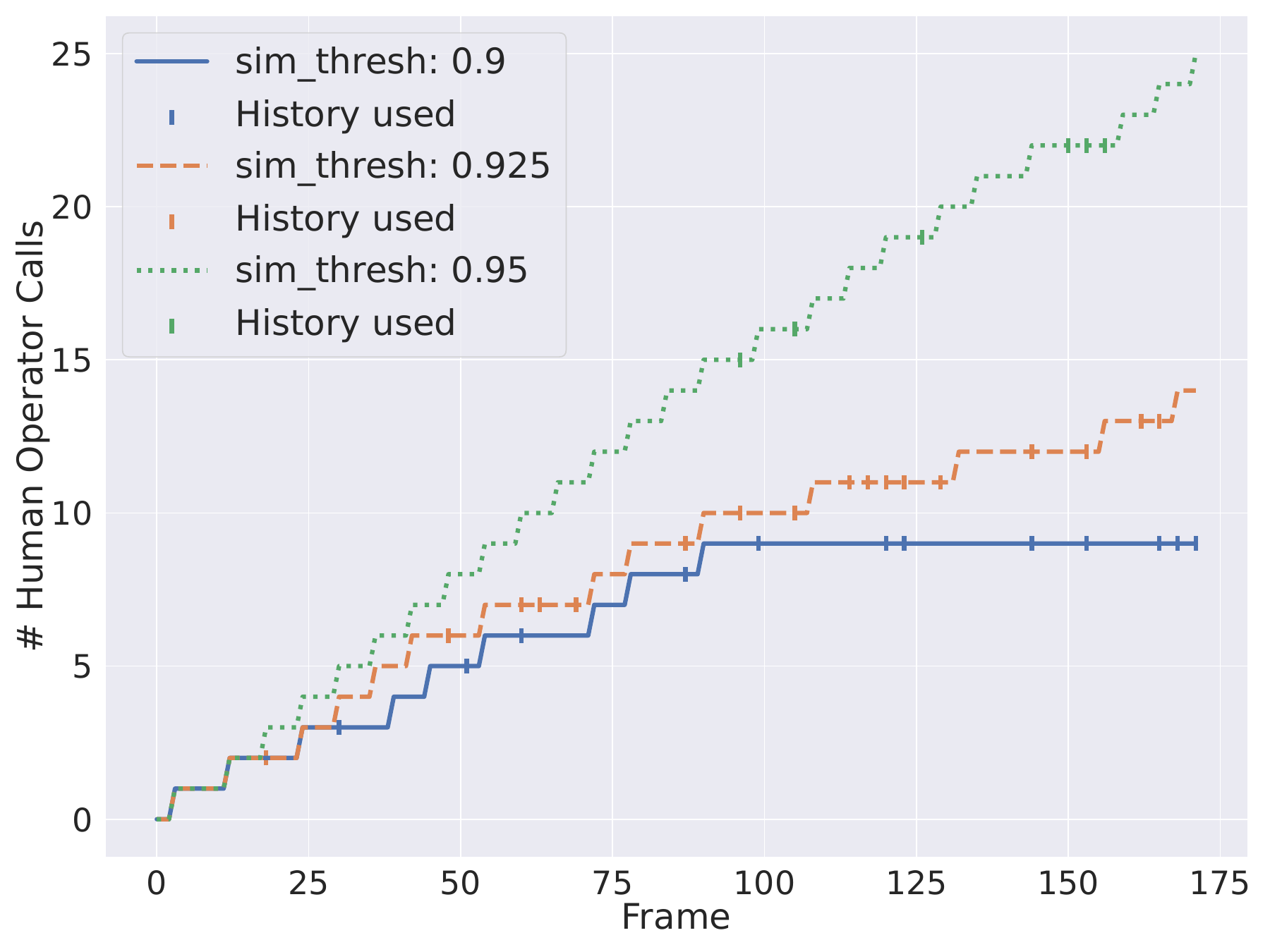}
            \caption{Freiburg Forest (DeepScene)}
            \label{fig:hoc-deepscene}
        \end{subfigure}
        \begin{subfigure}[b]{0.49\columnwidth}
            \centering
            \includegraphics[width=\textwidth]{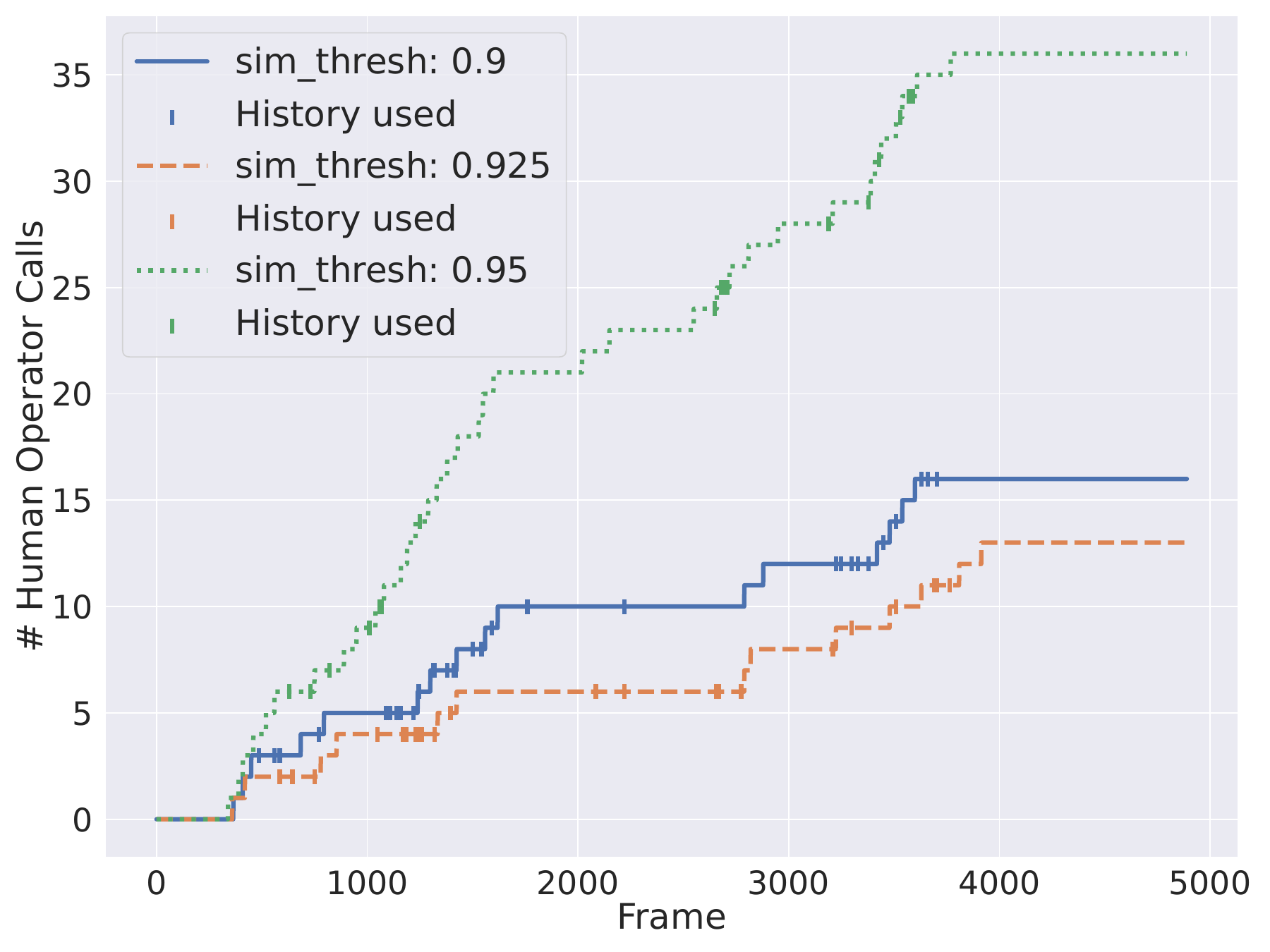}
            \caption{Forest Nursery (Ours)}
            \label{fig:hoc-ours}
        \end{subfigure}
        \caption{Human operator intervention frequency across different environments and similarity thresholds. The plots show cumulative operator calls progressing with frames, demonstrating how intervention requirements decrease as the system builds familiarity with the environment. Notably, after initial exposure, most environments show a significant plateau in new calls, with only occasional interventions for truly novel obstacles.}
        \label{fig:hoc-plots}
    \end{figure}
}

\subsection{Different Region of Interest for Different Vehicles} 
    \label{sec:roi-exp}
    \begin{figure}[ht]
        \centering
        \begin{subfigure}[b]{0.35\columnwidth}
            \centering
            \includegraphics[width=\textwidth]{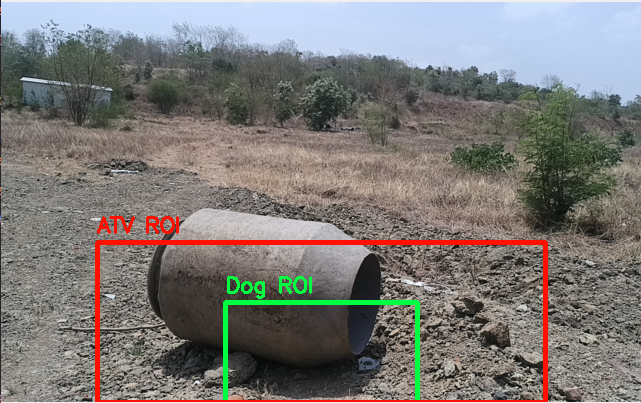}
            \caption{}
            \label{fig:roi-exp_atv-dog}
        \end{subfigure}
        \begin{subfigure}[b]{0.35\columnwidth}
            \centering
            \includegraphics[width=\textwidth]{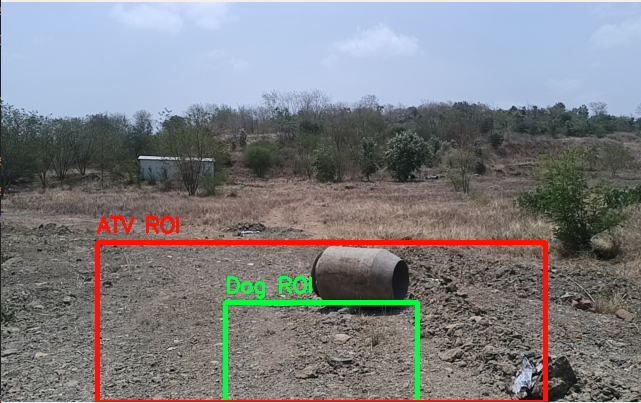}
            \caption{}
            \label{fig:roi-exp_atv-atv}
        \end{subfigure}
        \begin{subfigure}[b]{0.35\columnwidth}
            \centering
            \includegraphics[width=\textwidth]{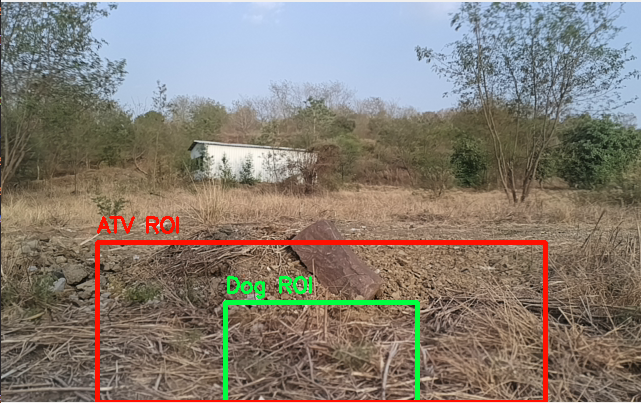}
            \caption{}
            \label{fig:roi-exp_dog-atv}
        \end{subfigure}
        \begin{subfigure}[b]{0.35\columnwidth}
            \centering
            \includegraphics[width=\textwidth]{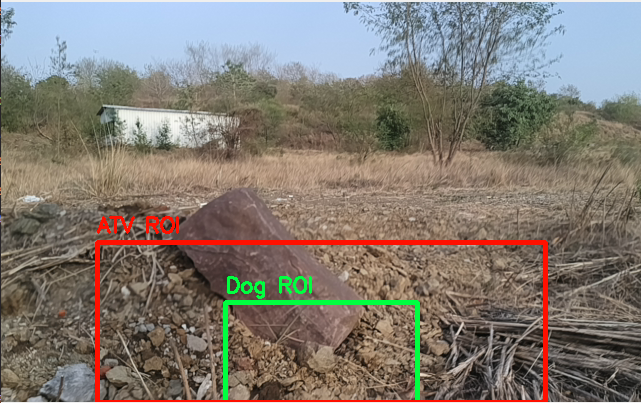}
            \caption{}
            \label{fig:roi-exp_dog-dog}
        \end{subfigure}
        \caption{Different vehicles (ATV, quadruped), detect unknown objects using different ROIs -- red for ATV, green for quadruped. Images (a) and (b) are from the ATV, (c) and (d) are from the quadruped}
        \label{fig:roi-exp}
    \vspace{-10pt}
    \end{figure}
    We conducted an experiment with two robots, the quadruped in \autoref{fig:robot-kombai}, and an all-terrain vehicle (ATV) \autoref{fig:robot-atv}. Both were provided with the prompts $\pi_1 = \text{``dirt path''},\, \pi_2 = \text{``dry grass''}$. Unknown objects were placed in their respective paths, namely a drum and a big rock. As discussed in \autoref{fig:roi-exp}, both robots have different ROIs. The aim of the experiment was to study the effect of exchanging the ROIs of these robots on the efficiency of \any in detecting unknown objects (See \autoref{fig:roi-exp}). If the ATV relies on the ROI of the quadruped, as shown in \autoref{fig:roi-exp_atv-dog}, the unknown object (drum) is detected too late to avoid, thus risking damage to the vehicle. Whereas if the ATV relied on the ATV's ROI, as shown in \autoref{fig:roi-exp_atv-atv}, it can detect the unknown object at sufficient distance to call the human operator to perform new actions. Similarly, if the quadruped relies on the ROI of the ATV, as shown in \autoref{fig:roi-exp_dog-atv}, then it detects the unknown object (rock) even when it is not in the quadruped's path. It makes an unnecessary call to the human operator, thus increasing their load. This issue is mitigated when using the ROI of the quadruped, in \autoref{fig:roi-exp_dog-dog}, as the unknown object is detected only when it is close enough to pose a risk of collision and damage, thus \any makes a sensible call to the human operator.

\section{Main Experimental Insights}
AnyTraverse achieved strong performance across all evaluated datasets with Mean Intersection over Union (MIoU) scores ranging from 0.8151 to 0.8521 (Table~\ref{tab:comparative}). These results demonstrate the framework's ability to generate accurate traversability masks across diverse off-road environments without requiring dataset-specific training. Our comparative analysis in Table~\ref{tab:comparative} shows that while specialized methods like OffSeg and GA-Nav achieve marginally higher performance on specific datasets, AnyTraverse maintains competitive performance (average MIoU of 0.825) while offering greater flexibility and adaptability. This is particularly significant considering that AnyTraverse does not require extensive training data or dataset-specific fine-tuning. Our key finding regarding human operator workload is that intervention requirements decrease significantly over time as the system builds familiarity with environments. \autoref{fig:hoc-plots} quantitatively demonstrates how calls due to unseen environments plateau, while ROI-based calls continue only for genuinely challenging terrain. At a similarity threshold of 0.925, the system achieved an appropriate balance between autonomy and safety, requiring operator input for approximately 15\% of frames in novel environments, dropping to below 5\% after environmental familiarity was established. This pattern was consistent across all tested datasets, with the most significant reductions observed in the RELLIS-3D and RUGD environments (Figures~\ref{fig:hoc-rellis} and~\ref{fig:hoc-rugd}).

The experiment described in \autoref{sec:roi-exp} shows the importance of proper choice of ROIs for different vehcicles, highlighting a trade-off between how early an unknown object should be detected and increased load on the human operator in these situations. The  faster a robot's charactersitic speed of traversal, and the bigger its size, the bigger the ROI should be.


\section{Conclusion}
In this work, we presented AnyTraverse, a zero-shot visual language model (VLM)-based segmentation framework for off-road traversability estimation with a human-in-the-loop design. By leveraging zero-shot learning, the framework exhibits strong adaptability to previously unseen environments, eliminating the need for extensive data collection and retraining. The inclusion of a human-in-the-loop further enhances the system’s robustness by enabling dynamic prompt generation for novel obstacles or drastically different terrains. Additionally, human-provided prompt weights allow the traversability estimation to be tailored to the specific requirements of different robotic platforms, ensuring greater operational flexibility and generalizability.
As part of future work, we aim to extend this framework by generating adaptable costmaps from the traversability segmentation outputs and integrating it with a navigation and control module for real-world autonomous deployment.

 \section*{Acknowledgement}
This work is partially supported by TiHAN project No. 384 and IISER Bhopal. 
\bibliographystyle{plain}
\bibliography{refs}
\end{document}